\documentclass[11pt,a4paper]{article}
\usepackage[hyperref]{acl2021}
\usepackage{times}
\usepackage{latexsym}
\usepackage{amssymb}
\usepackage{graphicx}
\usepackage{paralist}
\usepackage{graphicx}
\usepackage{subcaption}
\usepackage{caption}
\usepackage{multirow}
\usepackage{multicol}
\usepackage{booktabs}
\usepackage{array}
\newcolumntype{M}[1]{>{\centering\arraybackslash}m{#1}}

\usepackage{amsmath}
\DeclareMathOperator*{\argmax}{argmax} 

\newcommand*\samethanks[1][\value{footnote}]{\footnotemark[#1]}

\usepackage{microtype}

\aclfinalcopy 
\def\aclpaperid{2740} 

\title{Does Robustness Improve Fairness? Approaching Fairness with Word Substitution Robustness Methods for Text Classification}

\author{Yada Pruksachatkun \thanks{* Equal contribution} \\
  Amazon Alexa \\
  \texttt{ypruksac@amazon.com} \\\newline \\
  \textbf{Rahul Gupta} \\
  Amazon Alexa \\
  \texttt{gupra@amazon.com} \\\And
  Satyapriya Krishna \samethanks \\
  Amazon Alexa \\
  \texttt{satyapk@amazon.com} \\\And 
  \textbf{Jwala Dhamala} \\ 
  Amazon Alexa \\
  \texttt{jddhamal@amazon.com} \\\newline \\
  \textbf{Kai-Wei Chang} \\
  UCLA, Amazon Alexa \\
  \texttt{kwchang@cs.ucla.edu} \\}

\date{}

\begin{document}

\maketitle

\begin{abstract}
Existing bias mitigation methods to reduce disparities in model outcomes across cohorts have focused on data augmentation, debiasing model embeddings, or adding fairness-based optimization objectives during training. Separately, certified word substitution robustness methods have been developed to decrease the impact of spurious features and synonym substitutions on model predictions. While their end goals are different, they both aim to encourage models to make the same prediction for certain changes in the input. 
In this paper, we investigate the utility of certified word substitution robustness methods to improve equality of odds and equality of opportunity on multiple text classification tasks. We observe that certified robustness methods improve fairness, and using both robustness and bias mitigation methods in training results in an improvement in both fronts. 
 
\end{abstract}

\section{Introduction}
As natural language processing (NLP) technologies are  increasingly used in essential real-world applications, such as social media, healthcare, personal assistants and law \citep{he2020pathvqa,ahmad2020fairness}, it is important to ensure these systems do not create unintended outcomes for end-users or offer disparate experiences to customers from diverse backgrounds. This includes ensuring that model performance does not significantly differ across people belonging to different cohorts, such as different gender or race groups. 

A major subset of industry NLP applications lies in text classification, such as domain and intent classification in voice assistants \citep{Su2018ARS} or code tagging in healthcare \citep{Kemp2019ImprovedHP}. In this study, we focus on toxicity classification~\citep{dixon2018measuring} and occupation classification of Wikipedia biographies~\cite{DeArteaga2019BiasIB}. For toxicity classification, ensuring fairness means ensuring that a  model can identify toxicity to a similar accuracy across all examples regardless of the protected groups present in the example. Past studies (e.g,
\citep{dixon2018measuring,Zhang2020DemographicsSN,zhao2020logan}) have shown that toxicity classification models will falsely classify text containing certain protected attributes as toxic. Leading social media platforms and internet companies use toxicity classification models for content moderation \citep{Gorwa2020AlgorithmicCM}, thus having bias in such models can lead to increased silencing of under-served groups. Similarly, for occupation classification, a fair model should correctly identify occupations given a biography, regardless of the protected group that a person belongs to \cite{DeArteaga2019BiasIB}.

Recently, several studies have demonstrated societal bias in NLP systems~\citep{SocialBI,Tan2019AssessingSA,Liang2020TowardsDS} and various approaches have been proposed to mitigate the bias. These approaches include creating balanced datasets \citep{Park2018ReducingGB,zhao2018gender}, developing methods optimized for particular fairness notions \citep{zhang-etal-2017-adversarial,Zhang2020DemographicsSN}, model calibration~\cite{zhao2017men,jia2020mitigating}, and reducing representational bias \citep{bolukbasi2016man,zhao2018learning,Liang2020TowardsDS}. 

Separately, certified robustness approaches \citep{Jia2019CertifiedRT, Ye2020SAFERAS} have been developed to ensure robustness against word substitution attacks. Specifically, these strategies ensure small perturbations in the input embedding space do not alter model predictions. Despite never having been discussed in prior literature, this corresponds to notions of fairness, since protected attribute information (e.g. gender) is often irrelevant to the task at hand (i.e.  ``She is a good singer'' and  ``He is a good singer'' should have the same sentiment label). Thus, we posit that word substitution robustness methods can be used to make models invariant to protected attribute tokens and identifiers.

We explore the effect of robustness methods on fairness with GloVe-based CNN models \citep{Kim2014ConvolutionalNN} trained with Interval Bound Propagation (IBP) \citep{Jia2019CertifiedRT}, and BERT \citep{bert} trained with SAFER \citep{Ye2020SAFERAS}. We compare the effect of these robustness methods to popular bias mitigation methods. We find that robustness methods achieve promising performance on fairness metrics exceeding that of bias mitigation methods in several text classification tasks on gender and sexual orientation dimensions. Furthermore, training on both fairness and robustness exceeds performance over robustness and bias mitigation methods alone. Comprehensive analysis and visualization demonstrate that the robust methods decrease feature importance on gender tokens.

Our contributions are two-fold. First, we show that certified robustness methods can be used and integrated with bias mitigation methods to effectively improve models' performance on several notions of fairness, notably equality of opportunity and equality of odds. Secondly, by integrating robustness methods with fairness, we can improve a model's robustness while reducing bias, which is important in creating trustworthy NLP systems. Our study's practical implications include applications to models used in the industry that can handle customer inputs that may differ from the training data (robust) and that minimize any unintended consequences on the customers (fair). With this study, we aim to motivate future work geared towards developing methods that jointly optimize for multiple trustworthy aspects of models; specifically, those addressing model robustness and fairness.



\section{Mitigating Bias through Certified Robustness Methods}
\label{sec:fairrobust}

In the following, we first define the notions of fairness considered in this paper. Then, we discuss certified robustness methods, and how they can be applied to reduce bias in models. 

\subsection{Fairness Notions}
\label{sec:fairness}
We focus on measuring two notions of fairness in this paper -- equalized odds and equality of opportunity, as they are commonly used in quantifying bias in NLP applications. We describe the metrics associated with these notions in Section \ref{sec:methodology}. We give application examples of these notions on toxicity and occupation classification for English texts.


\paragraph{Equalized Odds}
 A model achieves \emph{Equalized Odds} \citep{Hardt2016EqualityOO} with respect to a protected attribute $A$ and outcome $Y$ if 
  $P(\bar{Y} = 1 | A = 0, Y=y) = P(\bar{Y} = 1| A = 1, Y=y)$, for $y \in \{0,1\}$. Protected attributes are traits or characteristics that cannot be discriminated against by law\footnote{\url{https://www.eeoc.gov/employers/small-business/3-who-protected-employment-discrimination}}. Intuitively, this means that the model should have equal true positive and false positive rates across groups. For toxicity classification, equalized odds implies that a model should be able to effectively detect toxicity on comments that include identifiers across all protected attribute cohorts, while not silencing any one cohort. Prior studies demonstrate that models disproportionately predict sentences associated with LGBTQ individuals as toxic, which may further silence discussion around LGBTQ issues and the voices of LGBTQ people \citep{oliva2020fighting}.

\paragraph{Equality of Opportunity} A model achieves \emph{Equality of Opportunity} \citep{Hardt2016EqualityOO} with respect to a protected attribute $A$ and outcome $Y$ if $P(\bar{Y} = 1 | A = 0, Y=1) = P(\bar{Y} = 1| A = 1, Y=1)$. This is a relaxation of Equalized Odds to the positive outcome, in which the model must have equal true positive rates across groups. For occupation classification, equality of opportunity implies that a model is able to correctly classify biographies of people from all groups, thus enabling equity in positive outcomes such as appropriate and useful matches in job recommendation sites.

Due to bias in the training data, off-the-shelf models often contain biases and disparities in model performance against underrepresented groups. Various bias mitigation approaches have been proposed to ensure the fairness in model predictions. We include a diverse array of bias mitigation methods, spanning embedding debiasing, in-training, and post-processing, as baselines. These consist of instance weighting \citep{Zhang2020DemographicsSN}, \textit{HardDebias} word embeddings \citep{Bolukbasi2016ManIT}, and adversarial debiasing \citep{Zhang2018MitigatingUB}. See more discussion in Sec. \ref{sec:methodology}.


\subsection{Certified Robustness for Bias Mitigation}
Designed for a different purpose, certified robustness methods present ways to train models that satisfy guarantees of word substitution robustness. By adapting certified robustness methods to fairness applications, we aim to make models invariant to spurious protected attribute information present in inputs, and thus improve in equality of opportunity and equalized odds.

Formally, a model \textit{f} is \textit{certifiably robust} if, for any example sentence $x$, and sentences $x'$ that consist of $x$ modified with word substitutions, $f(x) = f(x') = y$. In the robustness context, word substitution consists of swapping a word with its synonyms (usually defined using retrofitted word embeddings). For example, if $x$ = ``The waiter talked to the customer about their problems,'' $x'$ may consist of the sentences ``The waitress talked to the customer about their qualms.'' In the context of fairness, we consider 'waiter' and 'waitress' or gender pronouns to carry the same meaning in the context of toxicity and occupation classification, and to have the same label.


In this paper, we use two recently developed certified robustness methods, Interval Bound Propagation (IBP) \cite{Jia2019CertifiedRT} and SAFER \cite{Ye2020SAFERAS}. Given a set of perturbations for each word, these two models  ensures that word substitution do not affect model predictions. In particular, for each word, and a polytope spanned by the potential substitutions for that word in the embedding space, these methods ensure that swapping the word with any point in the polytope will not change the model predictions. To accomplish this, IBP minimizes the upper bound of the set of losses over perturbation sets, and SAFER uses a model-agnostic randomized smoothing technique. 

Both IBP and SAFER encourage models to be robust to spurious word substitutions, which include tokens that contain protected attribute information. The perturbations included in the original paper from \citet{alzantot} are based on a GloVe embedding that has been modified such that synonyms are close together. While the perturbation set does not include explicit gender and sexual orientation swaps (`boy' is not included in the perturbation set for 'girl', while 'girls', and 'women' are), we posit that certified robustness methods can still be applied to bias mitigation by improving robustness in examples that contain identifiers of underrepresented groups. Doing so will decrease model performance disparity in underrepresented group cohorts, and thus fulfill fairness notions.

\paragraph{IBP~\cite{Jia2019CertifiedRT}} IBP computes bounds on the model loss based on bounds on the input. The robustness goal of the IBP method is to minimize $\max F(x, \theta)$.
Here, $F(x, \theta)$ denotes the set of losses of a model over $B_{perturb}$, where $B_{perturb}$ is the set of perturbations for an example sentence $x$. Formally, $ F(x, \theta) = (f(\bar{x}, \theta) | \bar{x} \in B_{perturb})$.
The full loss for IBP is $ (1 - \lambda)f(x, \theta) + \lambda \mu^{final}(x, \theta),$ where $\mu^{final}$ is the upper bound on the loss $f(x, \theta)$ and $\lambda \in [0,1]$.

\paragraph{SAFER~\citep{Ye2020SAFERAS}}  Unlike IBP, SAFER does not require any changes to the model training. Instead, it employs a randomized smoothing mechanism in which an input is perturbed before being fed to the model during the training time. Specifically, SAFER creates random word substitutions using a perturbation set derived from a synonym network. 
 \citet{Ye2020SAFERAS} determine certified robustness of a model on an example by certifying that, given an example $z$, model score $s(z)$, and $y_{B} = \argmax_{c \in Y, c \neq y}s(z)$, the model score of the gold label $y$ is higher than the model score of the highest scoring non-gold label $y_B$ by a constant.

\section{Empirical Study on the Connection between Fairness and Robustness}
\label{sec:methodology}
To better understand the connection between fairness and certified robustness in the context of text classification, we empirically analyze models augmented with various combinations of robustness and fairness methods, as enumerated below.

\begin{compactenum}
  \item \textbf{Classifier (Baseline)}: The base text classification models. We consider two types of classification models that widely used in the literature, CNN \citep{Kim2014ConvolutionalNN} and BERT \citep{Devlin2019BERTPO}.
  \item \textbf{Classifier + Fairness}: Text classifiers trained with bias mitigation techniques (see Sec. \ref{sec:fairrobust}). 
  \item \textbf{Classifier + Robustness}: Text classifiers trained with robustness methods (see Sec. \ref{sec:fairrobust}).
  \item \textbf{Classifier + Robustness + Gender Word Perturbations}: 
To ensure that the model becomes robust against gender substitutions, we add definitional gender pairs (e.g., swapping he with she)~\cite{bolukbasi2016man} in the permutation set of IBP and SAFER.

  \item \textbf{Classifier + Robustness + Fairness}: Text classifier trained with both fairness and robustness objectives.
\end{compactenum}   

We aim to answer the following research questions based on the aforementioned configurations. (1) \emph{What is the effect of robustness methods on mitigating bias} (compare configuration 3 with 1 and 2)? (2) \emph{What is the effect of adding gender word substitutions to the robustness perturbation sets} (compare configurations 3 and 4)? (3) \emph{What is the effect of integrating bias mitigation and robustness methods} (compare configuration 5 with 1 and 3)?.
 
In particular, to answer the last question, we consider combining popular bias mitigation approaches with IBP as follows.  
  \begin{compactitem}
\item \textbf{Debiased Word Embeddings + IBP}: We replace the GloVe embeddings in the baseline CNN model with the \textit{HardDebias} embeddings obtained from \cite{bolukbasi2016man}, while keeping the rest of the IBP training methodology the same.

\item \textbf{Instance weighting + IBP}:
We add the instance weights to each sample in the loss computation during IBP training. 

\item \textbf{Adversarial Training + IBP}: 
We perform multitask training, alternating between optimizing for robustness loss and adversarial debiasing loss.
We initialized our adversarial training with the IBP-trained model. 

\end{compactitem}
\begin{table*}[h!]
\small
\centering
\begin{tabular}{p{4.5cm}p{2cm}p{1.5cm}p{1.5cm}p{1.5cm}p{2cm}}
\toprule 
\multirow{2}{*}{Model} & {Raw task ($\uparrow$)} & \multicolumn{3}{c}{{Fairness ($\downarrow$) }} & {Robustness ($\uparrow$)}  \\
\cmidrule(lr){2-2}\cmidrule(lr){3-5} \cmidrule(lr){6-6}
  &  {AUC}&  {EOdds} & {FPED} & {TPED} & {CRA} \\ 
  \midrule
Baseline & \textbf{0.957} & 0.508 & 0.197  & 0.311  & 0.270   \\ 
\midrule
 IBP &  0.913 & 0.184 &  0.005 & 0.179  & \textbf{0.934}   \\
 $\text{IBP}_\text{gender}$ &  0.947 &  0.237 &  0.062 & 0.175   & 0.912 \\
  \midrule
 Instance weighting &  0.955 & 0.505 & 0.196 & 0.309 & 0.214  \\
  \textit{HardDebias}  & 0.951 & 0.525& 0.221 &  0.304   & 0.404  \\
   Adversarial Training& 0.955 & 0.491 & 0.198 & 0.293  &  0.644   \\
  \midrule
 IBP + Instance weighting& 0.889 & \textbf{0.165} &  \textbf{0.002}  & \textbf{0.163}  & 0.942  \\
IBP + \textit{HardDebias}   & 0.923 &  0.459 &  0.169 &  0.290  & 0.890   \\
IBP + Adversarial Training & 0.920  & 0.473 & 0.192  &  0.281  & 0.901   \\

\bottomrule
\end{tabular}
\caption{\label{font-table} Certified robustness and bias mitigation methods with CNN on Jigsaw dataset. 
The best performance for each column is boldfaced. Results show that the certified robustness method (IBP) improves both robustness and fairness with performance drops on the raw task accuracy. }

\label{table:jigsawresults}
\end{table*}

\begin{table*}[h!]
\centering \small
\begin{tabular}{p{4.5cm}p{2cm}p{2cm}p{2cm}}
\toprule 
\multirow{2}{*}{Model} & {Raw task ($\uparrow$)} & {Fairness  ($\downarrow$) } & {Robustness ($\uparrow$)}  \\
\cmidrule(lr){2-2}\cmidrule(lr){3-3} \cmidrule(lr){4-4}
  &  {AUC}  & { TPED } & {CRA} \\ 
  \midrule
Baseline  & \textbf{0.787} &  0.131  & 0.115   \\
  \midrule
IBP &  0.743 & 0.127  & 0.702  \\
 $\text{IBP}_\text{gender}$  & 0.749 & 0.104 & 0.711   \\
  
 \midrule

Instance weighting & 0.755 & 0.118 &  0.095   \\

\textit{HardDebias} &  0.767  &  0.106  & 0.070  \\

Adversarial Training  &0.773 & 0.114 & 0.180   \\
\midrule

IBP + Instance weighting & 0.732  &  0.113   & \textbf{0.719}  \\

IBP + \textit{HardDebias} &  0.735 & \textbf{0.101}  &  0.715 \\

 IBP + Adversarial Training & 0.725 &  0.112  & 0.693   \\
\bottomrule
\end{tabular}
\caption{\label{font-table} Experiment results on CNN models on the Bias in Bios dataset. We see that our best performing model consists of initiating IBP training with $HardDebias$ embeddings. }
\label{table:bibibpresults}
\end{table*}

\paragraph{Datasets} 
We use the following two text classification datasets to validate our hypothesis on different data distributions. 

\begin{compactitem}

    \item \emph{Jigsaw Toxicity}\footnote{The data is available at https://www.kaggle.com/c/jigsaw-unintended-bias-in-toxicity-classification} is a dataset for toxicity classification that consists of 1,804,874 training examples, which we split into train and validation sets of size 1,443,900 and 360,974 respectively. We take 97,320 examples from the public leaderboard as the test set. 
    
    \item \emph{Bias in Bios}~\citep{DeArteaga2019BiasIB} \footnote{The data is available at https://github.com/microsoft/biosbias} is a dataset for occupation classification derived from Common Crawl corpus. It consists of 178,619 train and 91,917 test examples.

\end{compactitem}

\paragraph{Evaluation Metrics} We evaluate models on three dimensions: (1) raw task performance, (2) model fairness, and (3) model robustness. For the raw task performance, we follow prior work in using accuracy and area under the ROC curve (AUC) to evaluate the performance of a model on the Bias in Bios dataset and the Jigsaw Toxicity dataset respectively. To measure the robustness of a model, we follow \citet{Jia2019CertifiedRT} and \citet{Ye2020SAFERAS} to use the certified robustness accuracy (CRA). For fairness, we follow the discussion in Section \ref{sec:fairness} to evaluate a model based on \emph{equalized odds} and \emph{equal opportunity}.

For fairness, we measure two metrics - i.e, True Positive Equality Difference (TPED) and  False Positive Equality Difference (FPED). The FPED and TPED is calculated as:
$$\sum_{z \in Z} |f_z - f_{overall}|,$$
where $f$ is FPR or TPR depending on whether we are computing FPED or TPED, and $Z$ refers to the set of all classes in a protected group. 
Note that TPED and FPED metrics do not take into account how well the model does - for example, a model that achieves a true positive rate of 0.0 for all groups will still have a TPED of 0.

 For Bias in Bios dataset, we chose equality of opportunity to measure fairness, since it is important to ensure job candidates are matched with job recommendations that are relevant to them. Since equality of opportunity necessitates equality in true positive rates across cohorts, we use TPED as the fairness evaluation metric for Bias in Bios. For Jigsaw Toxicity, we define fairness by equalized odds, since it is important for toxicity classifiers to be able to detect toxicity in content containing identifiers across all groups, while not silencing any one. The combination of FPED with TPED aligns with the \textit{Equalized Odds} definition of fairness \cite{borkan2019nuanced}, thus we define a score \textit{EOdds} as FPED + TPED for ease of analysis. Equalized odds is satisfied when FPED = 0 and TPED = 0, and thus when \textit{EOdds} = 0.

 For the scope of this paper and the limitations of the dataset, we study binary gender for Bias in Bios, and both gender (male, female, transgender, and non-binary) and sexual orientation (homosexual/straight, heterosexual, gay, lesbian, bisexual) for Jigsaw Toxicity classification. While we acknowledge that there are a multitude of important attributes, we constrain the scope of this study to the attributes present in text classification datasets.

\paragraph{Experiment Details}
All the experiments were conducted on p3dn.24xlarge and p3.2xlarge AWS compute nodes.\footnote{https://aws.amazon.com/ec2/instance-types/} 
The IBP runs took 48 hours for Jigsaw Toxicity and 34 hours for Bias in Bios, while SAFER took 53 hours with evaluation for Jigsaw Toxicity and 37 hours for Bias in Bios.

For the experiments with CNN, we follow \citet{Jia2019CertifiedRT} to configure the IBP schedule and CNN models. In particular, we used a CNN model with a hidden size of 100 and kernel size of 3 with the GloVe embedding~\cite{pennington-etal-2014-glove} as inputs. For IBP, we linearly increased the weight on the certified robustness objective from 0 to 0.8 for 40 epochs, before training for 20 epochs on the full certified robustness objective. 

For the experiments with BERT, we follow \citet{Ye2020SAFERAS} to configure the BERT model and SAFER experiment. We use \texttt{bert-base-uncased}, and take the top-100 words that are closest in cosine similarity for each token as the token's perturbation set.
We describe the remaining hyper-parameter details (learning rate, epochs, dropout probability) in the the appendix, which we obtained after a hyper-parameter search on the development set.

\begin{table*}[]
\centering
\small
\begin{tabular}{p{3cm}p{2cm}p{1.5cm} p{1.5cm} p{1.5cm} p{2cm}}
\toprule 
\multirow{2}{*}{Model} & {Raw task ($\uparrow$)} & \multicolumn{3}{c}{{Fairness ($\downarrow$)}} & {Robustness ($\uparrow$)}  \\
\cmidrule(lr){2-2}\cmidrule(lr){3-5} \cmidrule(lr){6-6}
  &  {AUC}&  {EOdds} & {FPED} & {TPED} & {CRA} \\ 
  \midrule
Baseline & 0.914 & 0.553 & 0.290 &  0.263 & 0.950  \\
 SAFER & 0.918&  \textbf{0.286} & \textbf{0.144} & \textbf{0.142} & \textbf{0.967}   \\
 $\text{SAFER}_\text{gender}$ & \textbf{0.968} &  0.347 & 0.176 &  0.171 &  0.917  \\
\bottomrule
\end{tabular}
\caption{\label{font-table} Model performance on BERT (Baseline) and SAFER on the Jigsaw dataset. Similar to the observation with IBP, SAFER improves both the fairness and robustness metrics. }
\label{table:jigsawsaferresults}
\end{table*}

\begin{table*}[]
\centering
\small
\begin{tabular}{p{3cm}p{2cm}p{2cm} p{2cm}}
\toprule
\multirow{2}{*}{Model} & {Raw task ($\uparrow$)} & {Fairness  ($\downarrow$) } & {Robustness ($\uparrow$)} \\
\cmidrule(lr){2-2}\cmidrule(lr){3-3} \cmidrule(lr){4-4} 
 &  {AUC} & {TPED} & {CRA}\\ 
\midrule
Baseline & \textbf{0.796} &  0.148 & 0.164   \\
SAFER &  0.744  & 0.134 & 0.726  \\
SAFER$_{gender}$ &  0.761 &  \textbf{0.097} & \textbf{0.733}  \\
\bottomrule
\end{tabular}
\caption{\label{font-table} Model performance on BERT (baseline) with SAFER on the Bias in Bios dataset.}
\label{table:bibsaferresults}
\end{table*}

\section{Results}

The results for Jigsaw Toxicity and Bias in Bios are in Tables \ref{table:jigsawresults}, \ref{table:bibibpresults}, \ref{table:jigsawsaferresults} and \ref{table:bibsaferresults}.

\paragraph{Effect of certified robustness methods for mitigating bias}
We observe that adding IBP during training achieves better performance on fairness over othe bias mitigation approaches across Jigsaw Toxicity and Bias in Bios. In Jigsaw Toxicity, EOdds improves from 0.508 to 0.184, and in Bias in Bios, TPED improves from 0.131 to 0.127. Similarly, training models with SAFER results in an improvement in performance in all fairness metrics, with an improvement in EOdds from 0.553 to 0.286 in Jigsaw Toxicity and an improvement in TPED from 0.148 to 0.134 in Bias in Bios.

\paragraph{Effect of adding gender word substitutions to the robustness perturbation sets} While adding gender word substitutions further improves fairness in Bias in Bios, it results in worse fairness scores in Jigsaw Toxicity than plain certified robustness methods. In Bias in Bios, $\text{IBP}_\textit{gender}$ results in a lower TPED than all fairness only baselines. This trend holds in SAFER, where  $\text{SAFER}_\textit{gender}$ achieves lower TPED than SAFER. In Jigsaw Toxicity, adding gender words to the perturbation set degrades performance in equalized odds for both IBP and SAFER. This may be because the list of gender word substitutions do not include words relating to sexual orientation and non-binary gender, and thus may only improve fairness amongst examples containing male and female identifiers. 

\paragraph{Effect of integrating bias mitigation methods with certified robustness methods} Training model with both IBP and bias mitigation methods improves fairness metrics over fairness-only baselines in both datasets. In Bias in Bios, the model that comes closest to fulfilling equality of opportunity is the one trained with both IBP and $HardDebias$, which achieves a TPED of 0.101. In Jigsaw Toxicity, we see a similar trend, with improvements in EOdds after adding IBP training to instance weighting, $HardDebias$, and adversarial training. The model trained with both IBP and instance weighting achieves a EOdds score of 0.165, which is the lowest among all approaches

We also note that for Jigsaw Toxicity, instance weighting mitigates bias more effectively than $HardDebias$ and adversarial training (both in isolation and in combination with robustness methods). This is not the case for Bias in Bios, where $HardDebias$ and adversarial training is more effective than instance weighting in mitigating bias. This may be due to the fact that instance weighting mitigates bias explicitly for a wider array of sexual orientations and gender demographics than the other two methods. The original $HardDebias$ method only projects away the gender direction from embeddings. For adversarial debiasing, we train the adversary with the subset of the training set that is annotated for the presence of protected attribute groups, which is highly skewed towards male and female. Thus, $HardDebias$ and adversarial training may mitigate bias for binary gender, but fall short in mitigating bias for non-binary gender and sexual orientations. Conversely, instance weighting, which mitigates bias for a wider array of demographics, does not mitigate bias on gender in Bias in Bios as well as the other methods.

\paragraph{Additional Observations} Outside of the effects of robustness on fairness, we observe differing effects of the methods on certified robustness and raw accuracy. As expected, IBP and SAFER improves performance on certified robustness on both datasets. However, we also observe degradataions in raw task accuracy in experiments with robustness methods. Combining robustness with bias mitigation methods results in a degradataion of raw task performance over fairness-only baselines. Additionally, fairness-only training results in differing effects on certified robustness accuracy. Adversarial debiasing improves certified accuracy in both datasets, while \textit{HardDebias} embedding-initiated training results in an increase in certified robustness in Jigsaw Toxicity, but a decrease in Bias in Bios. This difference in findings may be due to the shorter length of examples in Jigsaw Toxicity, which has a median length of 34, compared with the median length of 72 in Bias in Bios, which in turn determines the number of possible perturbations used to calculate certified accuracy and the difficulty in achieving high CRA.

\begin{figure*}[h!]
\centering
    \includegraphics[width=0.8\textwidth]{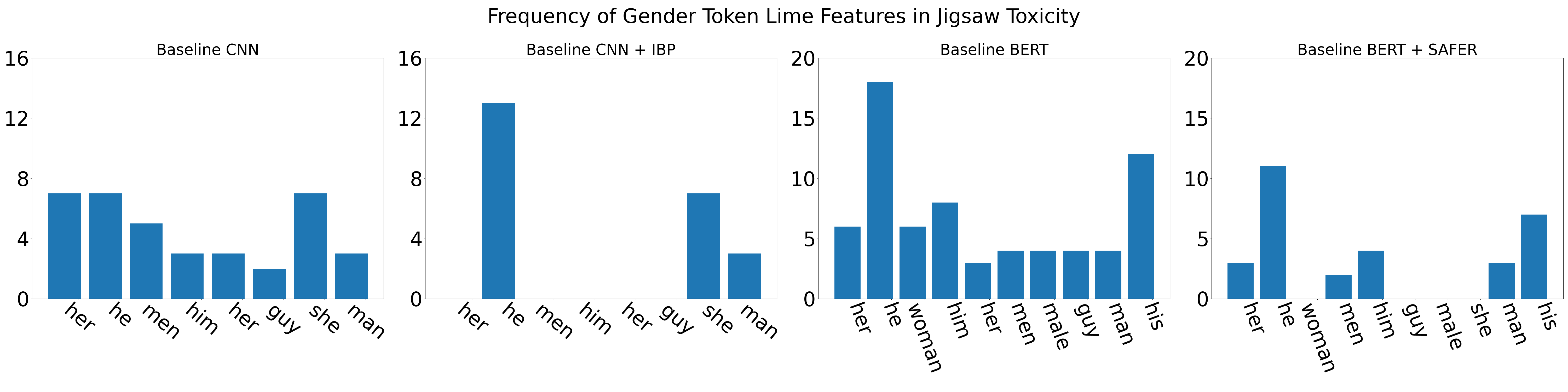}

    \includegraphics[width=0.8\textwidth]{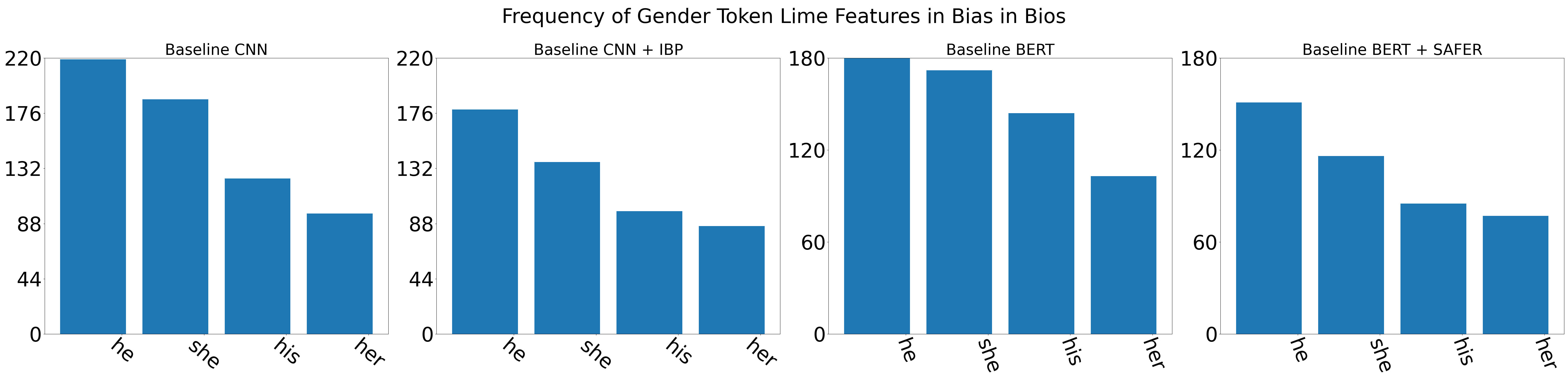}
    \caption{Frequency of gender token features as extracted by LIME for baseline and IBP trained models for Jigsaw Toxicity and Bias in Bios. We see a decrease in number and frequency of gender tokens in the list of top-5 (for Jigsaw Toxicity) and top-50 (for Bias in Bios) most important features.}
  \label{fig:bias_lime}
\end{figure*}

\section{Analysis}
In this section, we study how robustness training affects the features our models use for classification. We posit that robustness training encourages  models to focus more on predictive attributes than on protected attributes. To gain insight into this, we use LIME \citep{Ribeiro2016WhySI} on the baseline, IBP, and SAFER trained models and extract token features importance as assigned by the model. We run LIME on the subset $E$ of examples that are misclassified by our baseline model as toxic that are correctly classified by the IBP model. We take the top $k$ features for each of the examples (where $k$ = 5 for Jigsaw and $k$ = 50 for Bias in Bios) over $E$, and then count the number of gender tokens that appear in that list. For Jigsaw Toxicity, the number of examples that we run LIME on is 488 for CNN experiments and 182 for BERT experiments. For Bias in Bios, we run LIME over a random subset of 500 examples from $E$ for both CNN and BERT experiments.

For Jigsaw Toxicity, we see from Figure~\ref{fig:bias_lime} that LIME extracts less gender tokens in the top-5 features of the IBP-trained and SAFER-trained model compared to the baseline model. Notably, there are 37 gender tokens that appear in the CNN model, while only 23 in the IBP-trained model. Similarly, 69 gender tokens appear in the baseline BERT model while only 37 appear in the SAFER-trained one. For Bias in Bios, we see a similar trend from Figure~\ref{fig:bias_lime}. The number of important gender token features decreases from 626 to 500 after IBP training, and from 626 to 429 after SAFER. 

In addition, we compute the gradient with the output with respect to the input on several examples from Jigsaw Toxicity, which is shown in Table \ref{table:saliency}. We observe that the baseline model focuses on tokens related to protected groups, while the IBP model takes into account all parts of the sentence.

\begin{table*}[h!]
\centering 
\small
\begin{tabular}{|M{1.5cm}|M{13cm}|}
\hline {Model} & {Saliency Map} \\ \hline
\multicolumn{2}{|c|}{Example 1}\\ \hline
Baseline &  \includegraphics[width=0.8\textwidth]{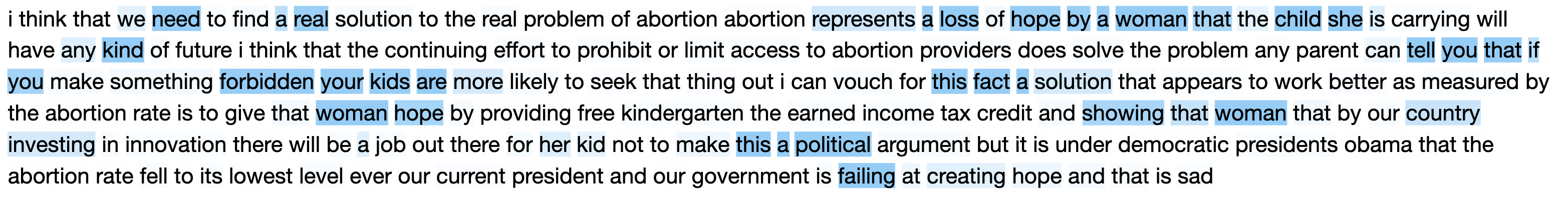} \\
\hline
IBP & \includegraphics[width=0.8\textwidth]{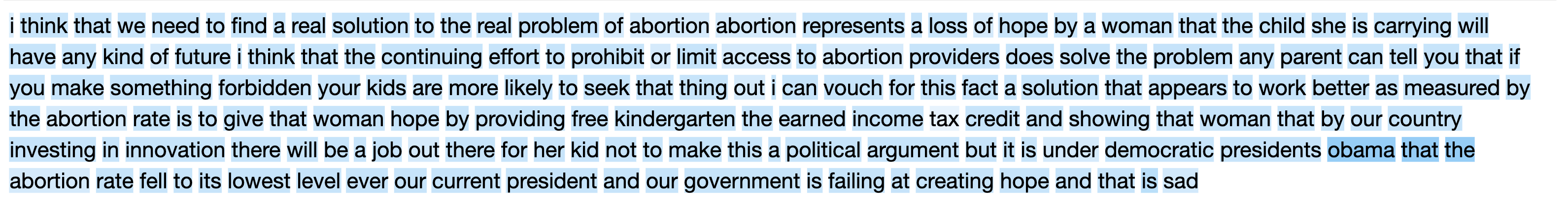} \\
\hline
\multicolumn{2}{|c|}{Example 2}\\  \hline 
Baseline & \includegraphics[width=0.8\textwidth]{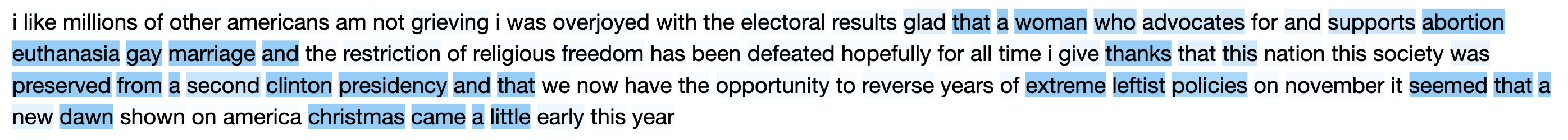} \\
\hline
IBP & \includegraphics[width=0.8\textwidth]{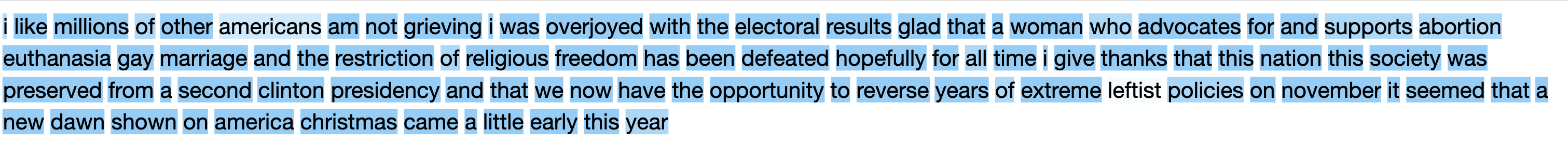} \\
\hline
\end{tabular}
\caption{Gradient saliency examples on Jigsaw Toxicity. Highlights show larger value of the output gradient with respect to the token embedding. The baseline CNN model focuses on some tokens related to protected groups (e.g., woman), while IBP encourages the model to take into account other parts of the sentence, resulting in less bias.} 
\label{table:saliency}
\end{table*}

\section{Related Work}
Much work has been done in studying fairness in various NLP models~\citep{Mehrabi2019ASO,Sun2019MitigatingGB,blodgett2020language}. In toxicity classification, \citet{Adragna2020FairnessAR} and \cite{Zhang2020DemographicsSN} study the fairness in predicting toxic internet contents in which the contents contain demographic identity-terms (e.g., ``gay", ``black"). In occupation classification,  \citet{DeArteaga2019BiasIB} and \citet{Romanov2019WhatsIA} study the impact of including explicit gender indicators such as a person's names or a pronoun in online biographies.

Some notable bias mitigation methods, which we also use in this paper, include instance weighting \citep{Zhang2020DemographicsSN}, embedding debiasing \citep{Bolukbasi2016ManIT,Wang2020DoubleHardDT}, and adversarial debiasing \citep{Zhang2018MitigatingUB}. 
In particular,  \citet{Bolukbasi2016ManIT} proposed to reduce representational harm existent in word embeddings. \citet{Zhang2020DemographicsSN} proposed instance weighting, a method to debias text classification models for bias against examples containing demographic identity-terms by weighting the instances in the loss function, and that is optimized for demographic pairty. 
\citet{Zhang2018MitigatingUB} presents an adversarial training approach to achieve various notions of fairness that is achieved by training an adversary to identify information on protected groups and training the model to minimize the adversary loss. These approaches are designed for reducing specific types of bias exhibited in data. 

On the robustness front, it has been shown that models are susceptible to adversarial word substitution attacks \cite{Ebrahimi2018HotFlipWA,Jia2017AdversarialEF}. Parallel to the development of methods developed to reduce word substitution robustness in the NLP domain (e.g., \cite{iclr-17:Miyato,huang2019achieving,zhou2021defense}), many studies has been done in the computer vision domain to ensure that models are robust to image noising~\citep{Kannan2018AdversarialLP,Szegedy2014IntriguingPO}. 

In the intersection of area between fairness and robustness of model training, 
there is limited prior work in the NLP area. 
\citet{Nanda2020FairnessTR} investigate and define \textit{robustness bias}, a notion of fairness in which a model must be impervious to perturbations to the same degree for all subgroups, and investigate robustness bias in the  computer vision domain. \citet{Adragna2020FairnessAR} examine the use of invariant risk minimization in improving the fairness on out-of-distribution data for toxicity classification. Their robustness approach is inspired from domain generalization and it allows to learn models that have invariant performance across different label distributions. 
This differs from the word substitution notions of robustness that our methods are optimized for.  \citet{Chang2020OnAB} shows that achieving equalized odds is incongruent with adversarial robustness on the COMPAS \citep{COMPAS} and the Adult dataset \citep{ADULT}, which is outside the NLP domain. The closest work to ours is in counterfactual logit pairing \citep{Garg2019Counterfactual}, which encourages a model to be robust to protected attributes for counterfactual fairness. However, logit pairing does have the certified characteristic of the robustness methods we use in this study.

\section{Conclusion}
We present a study that investigates the effect of optimizing for word substitution robustness on fairness. We find that, in both CNN and BERT models, adding robustness methods such as IBP and SAFER to the training process improves fairness metrics over adding bias mitigation methods alone. Given these promising results, we encourage future explorations in using robustness methods to not only improve fairness metrics, but to also optimize for both fairness and robustness, two important aspects of creating trustworthy NLP. 

 Future work may include studying the effects of robustness and fairness in attributes other than gender and sexual orientation, extending our study to other word substitution based robustness methods, and exploring more sophisticated methods to combine robustness and bias mitigation methods during training. We also intend on extending the study to investigating the impact of privacy preserving training methods on both, robustness and fairness.


\section*{Broader Impact}
We limit the scope of this paper to gender and sexual orientation in this initial effort, and future work must be done on mitigating bias in other protected attribute dimensions such as race, ethnicity, neurodiversity, etc. Additionally, this work draws importance to the need to extend fairness methods to groups beyond binary gender. In our $\text{IBP}_{\textit{gender}}$ experiments, we only consider swapping binary gender pairs from prior literature to provide an anchor for our analysis. We see from our results that methods that mitigate for binary gender such as \textit{HardDebias} and $\text{IBP}_\textit{gender}$ do not reduce harm for all gender or sexual orientation, especially for non-binary gender and non-heterosexual sexual orientation groups. We will extend the study in the future by developing fairness methods that directly mitigate for non-heterosexual sexual orientations and non-binary genders pairs using sociology literature.

The language used in this paper is English. We recognize that the presented methods rely on the availability and quality of the set of words associated to a fairness task. Scaling to languages beyond English--such as gendered languages like Spanish--need more careful analysis. Another limitation of this method is that word substitution may lead to non-sensible sentences and inappropriate grammar especially in complex fairness domains where it is difficult to find word-to-word mapping (e.g., mapping names of religious artifacts like Christmas tree or Diwali lights, etc are not trivial). 

Our experiments and results show that pursuing fairness can help in improving robustness and vice versa. With these findings, we hope to inspire researchers to investigate novel approaches that focus on jointly achieving robust and fair models. We also hope that this work will lead to more investigations around achieving multiple objectives such as privacy, robustness  and fairness together in the NLP research community. 

\bibliography{anthology,acl2020}
\bibliographystyle{acl_natbib}

\appendix

\section{Appendices}
 \textbf{Appendix A. Hyperparameter Settings}
 \label{app:hyper-parameters}
 We perform hyperparameter search on the dev set using random search with 12 trials, with initial learning rate range between $1 * 10^{-2}$ to $1 * 10^{-7}$, a dropout probability range of 0.1 to 0.5, and number of epochs between 10 and 60.The final hyperparameter settings are shown in Table \ref{table:hyp_ibp}. We choose our hyperparameters based on the one that minimizes FPED + TPED + (1 - CRA) + (1 - \textit{tp}), where \textit{tp} refers to task performance.

\begin{table*}[h!]
\centering
\begin{tabular}{|p{3cm}| p{2cm}|p{2cm}|p{2cm}| p{2cm}|p{2cm}|}
\hline \textbf{Experiment} &  \textbf{Learning Rate} & \textbf{Dropout Prob} & \textbf{Number of epochs}\\ \hline
GloVe + CNN (Jigsaw) & 1e-2 & 0.5 & 20  \\
\hline
GloVe + CNN (Bias in Bios) & 1e-3 & 0.1 & 15  \\
\hline
BERT + SAFER (Jigsaw) & 5e-6 & 0.1 & 20  \\
\hline
BERT + SAFER (Bias in Bios) & 1e-5 & 0.1 &  15 \\
\hline

\end{tabular}
\caption{\label{font-table} Hyperparameter settings for our experiments. We use the same hyperparameters across our fairness and robustness experiments. }
\label{table:hyp_ibp}
\end{table*}

Additionally, for adversarial debiasing, we tune the adversary loss weight from $\alpha=0.1$ to $\alpha=3$, and choose $\alpha=1$ for the weight. We pretrain our classifier and adversary for 2 epochs each.
\end{document}